\documentclass[10pt,twocolumn,letterpaper]{article}
\usepackage{iccv}              %

\definecolor{cvprblue}{rgb}{0.21,0.49,0.74}
\usepackage[pagebackref,breaklinks,colorlinks,allcolors=cvprblue]{hyperref}
\usepackage{multirow}
\usepackage{colortbl}
\usepackage{makecell}
\usepackage{arydshln}
\usepackage{bm}

\def\paperID{8114} %
\def\confName{ICCV}
\def\confYear{2025}
\usepackage{bbding}

\title{TinyViM: Frequency Decoupling for Tiny Hybrid Vision Mamba}

\author{Xiaowen Ma $^*$
\quad
Zhenliang Ni \thanks{Equal contribution, $^{\dagger}$ Corresponding author}
\quad
Xinghao Chen$^{\dagger}$ \\
Huawei Noah’s Ark Lab \\
{\tt\small xwma@zju.edu.cn, \{nizhenliang2, xinghao.chen\}@huawei.com}
}

\begin{document}
\maketitle
\begin{abstract}
Mamba has shown great potential for computer vision due to its linear complexity in modeling the global context with respect to the input length. However, existing lightweight Mamba-based backbones cannot demonstrate performance that matches Convolution or Transformer-based methods. By observing, we find that simply modifying the scanning path in the image domain is not conducive to fully exploiting the potential of vision Mamba.  
In this paper, we first perform comprehensive spectral and quantitative analyses, and verify that the Mamba block mainly models low-frequency information under Convolution-Mamba hybrid architecture. Based on the analyses, we introduce a novel Laplace mixer to decouple the features in terms of frequency and input only the low-frequency components into the Mamba block. 
In addition, considering the redundancy of the features and the different requirements for high-frequency details and low-frequency global information at different stages, we introduce a frequency ramp inception, i.e., gradually reduce the input dimensions of the high-frequency branches, so as to efficiently trade-off the high-frequency and low-frequency components at different layers. By integrating mobile-friendly convolution and efficient Laplace mixer, we build a series of tiny hybrid vision Mamba called TinyViM.
The proposed TinyViM achieves impressive performance on several downstream tasks including image classification, semantic segmentation, object detection and instance segmentation. In particular, TinyViM outperforms Convolution, Transformer and Mamba-based models with similar scales, and the throughput is about 2-3 times higher than that of other Mamba-based models. 
Code is available at \href{https://github.com/xwmaxwma/TinyViM}{https://github.com/xwmaxwma/TinyViM}.

\end{abstract}

\begin{figure}[t]
    \centering    \includegraphics[width=0.5\textwidth]{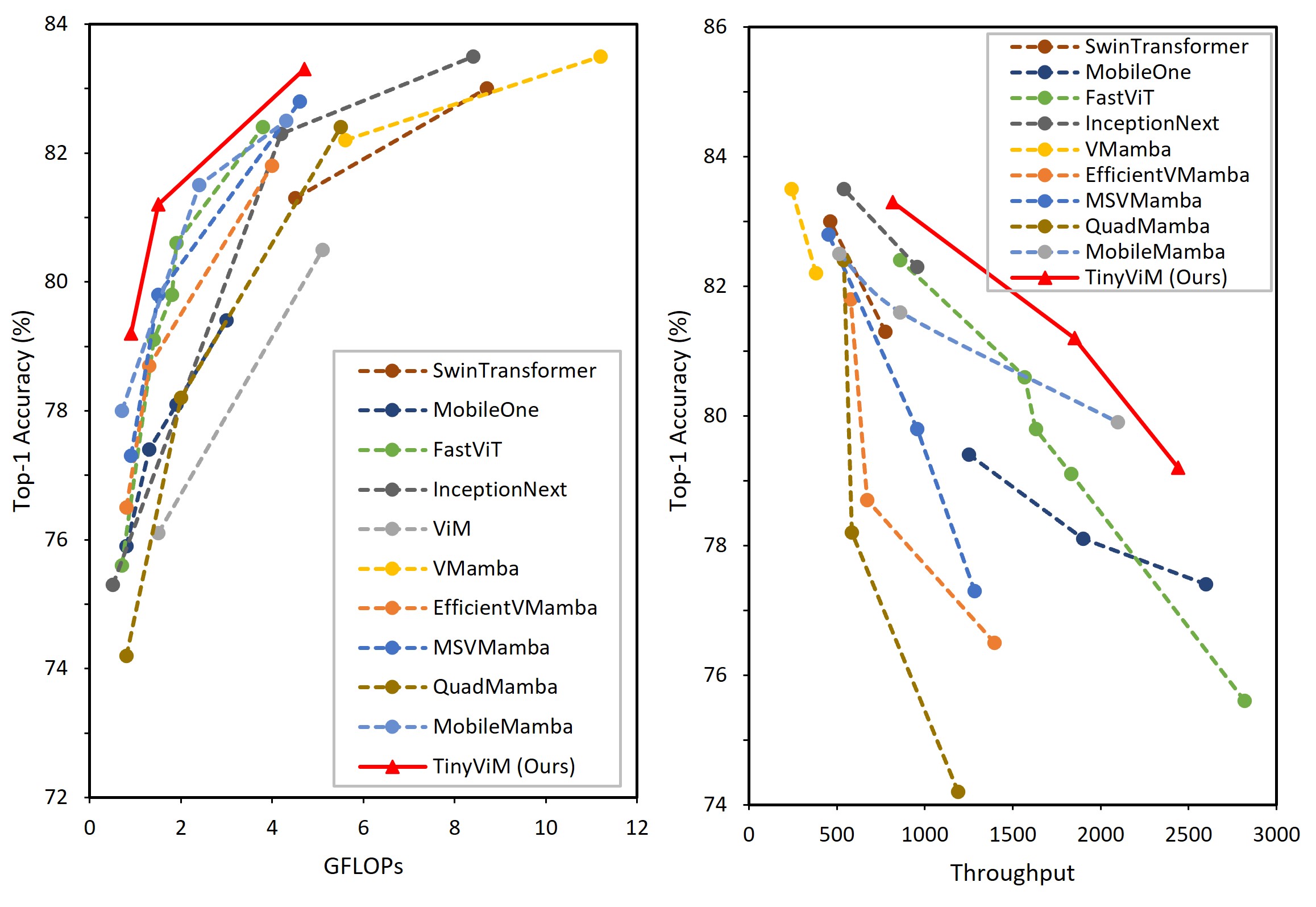}
    \caption{Comparision of GFLOPs, Throughput and accuracy between TinyViM and others. The top-1 accuracy is tested on ImageNet-1K and the throughput is measured on Nvidia V100 GPU with maximum power-of-two batch size that fits in memory. Compared to recent Mamba-based models such as EfficientVMamba, QuadMamba, and VMamba, TinyViM has higher top-1 accuracy with $2\times$ higher throughput and fewer GFLOPs.}
    \label{fig:comp}
\end{figure}
\section{Introduction}
\label{sec:intro}
\begin{figure*}[t]
    \centering    \includegraphics[width=0.99\textwidth]{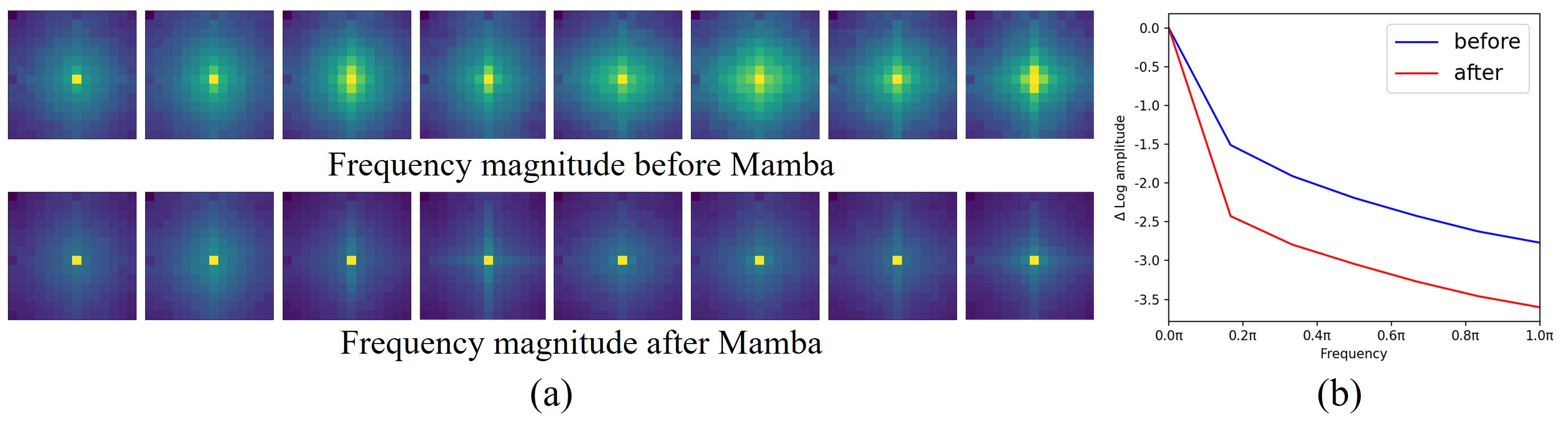}
    \caption{Spectral analysis of Mamba. (a) Frequency magnitude ($14 \times 14$) from 8 output channels before- and after Mamba block. (b) Relative log amplitudes of Fourier transformed feature maps. The magnitude and amplitude are averaged over 384 samples. (a) and (b) show that Mamba block focuses on capturing low-frequency information under the Convolution-Mamba hybrid architecture.}
    \label{fig:freq}
    \vspace{-2mm}

\end{figure*}
In recent years, Mamba has gained widespread attention and application in the field of vision due to its advantages in long-sequence modeling \cite{mamba}. Mamba is a selective structured state-space model that can overcome the limitations of local perception in Convolution neural networks \cite{resnet} and the quadratic computational complexity of Transformers \cite{vit}, thereby demonstrating strong capabilities in handling long-sequence data. Due to its advanced context extraction capabilities, Mamba has been widely applied in image classification tasks and has achieved advanced performance, such as Vim \cite{vim}, VMamba \cite{vmamba}, etc.

Vim~\cite{vim} is one of the earliest works to apply Mamba to visual tasks, by introducing a bidirectional state-space model to process 2D image data. By introducing the Cross-Scan Module (CSM) and 2D Selective Scanning (SS2D) modules, VMamba~\cite{vmamba} reduces the computational complexity from quadratic to linear while maintaining the global acceptance domain. MSVMamba~\cite{msvmamba} uses a multi-scale 2D scanning method to apply Mamba to both original and down-sampled feature maps. This not only helps to learn long-distance dependencies, but also reduces computational costs. EfficientVMamba~\cite{evmamba} is a lightweight Vision Mamba variant that reduces computational complexity through dilated-based selective scanning. 
Although these methods inherit the advantages of Mamba linear complexity and global receptive field, they do not show competitive performance with lightweight backbones based on Convolution \cite{mobileone, inceptionnext} and Transformer \cite{swiftformer,fastvit,agent} in computationally limited and real-time deployment scenarios, as shown in Fig.~\ref{fig:comp}. 
In fact, these methods that simply modify the scan path in the image domain do not utilize the full potential of visual Mamba \cite{vmamba, vim}. 
In this paper, we try to rethink the design of lightweight vision Mamba from the perspective of frequency decoupling, which is a reasonable and effective solution.

Through observation, we find that Mamba prefers to extract low-frequency features and ignores some high-frequency features. As shown in Fig.~\ref{fig:freq}, we construct a baseline using mobile-friendly Convolution and vanilla Mamba, and perform spectral analysis of the features before and after the Mamba. Many previous works~\cite{efficientformer,swiftformer,repvit} have verified the necessity of using Convolution in lightweight backbones. After applying the Mamba block, the low-frequency component in the center of the two-dimensional spectrum map is highlighted, and the high-frequency component is suppressed. Therefore, to improve the efficiency of low-frequency modeling and keep the high-frequency features, we introduce an efficient Laplace pyramid to decompose the high- and low-frequency components of the features.  
Then, we input only the low-frequency components into the Mamba block to obtain a global receptive field, and for the high-frequency components, we use a reparameterized 3$\times$3 depth-wise Convolution to enhance high frequency information \cite{conv1,conv2}. 
 It is worth noting that the size of the low frequency feature is much smaller than that of the original input feature, which can greatly reduce the computational complexity and improve the throughput of the Mamba. Therefore, the strategy of frequency decoupling and specifically enhancement separately effectively improves the efficiency and the feature representation ability of the model.

 In addition, previous models based on Convolution or Transformer have verified that: 1) deep neural networks have feature redundancy \cite{ghostnet}; 2) deep neural networks tend to require more high-frequency information in the shallow layer and more global information in the deep layer \cite{inceptiontransformer,scsm}. Therefore, we propose the frequency ramp inception to further improve the performance and efficiency of the model. Specifically, we split the feature map along the channels at each stage.
 Early stages prioritize high frequencies, reflected in a higher channel ratio for the high-frequency branch, while later stages concentrate on low frequencies, allowing for a greater channel allocation to low frequencies. This strategic distribution enables the network to learn features that are more suitable at each stage of processing effectively.

By integrating frequency decoupling and frequency ramp inception, we propose a tiny and efficient Convolution-Mamba hybrid architecture TinyVim. We then conduct extensive experiments to validate the effectiveness of TinyViM, which achieves state-of-the-art image classification performance on ImageNet and advanced performance in downstream tasks such as detection and segmentation. Specifically, TinyViM has higher classification accuracy than other Convolution or Transformer-based backbones with a similar scale. In addition, compared to other Mamba-based backbones such as EfficientVMamba-T, TinyViM has $2 \times$ higher throughput and 2.7\% higher accuracy, as shown in Fig.~\ref{fig:comp}. Our contributions can be summarized as follows.
\begin{itemize}
    \item We verify that Mamba mainly models low-frequency information under the Convolution-Mamba hybrid architecture based on qualitative analysis and quantitative comparison.
    \item We introduce an efficient Laplace mixer, which improves the modeling efficiency and performance of visual Mamba through frequency decoupling and frequency ramp inception.
    \item We propose TinyViM by integrating the Laplace mixer and Convolution. Extensive experiments show that TinyViM outperforms other Convolutional, Transformer and Mamba-based models with a similar scale, and achieves a better trade-off between performance and efficiency.
\end{itemize}

\section{Related Work}
\subsection{Efficient Generic Vision Backbones}
 Convolution has been widely employed for designing efficient vision backbones due to the inductive bias and efficient parallel processing on GPUs. Early works such as the MobileNet series \cite{mobilenet, mobilenetv2, mobilenetv3} propose the depth-wise separable Convolution and the inverted residual blocks to further reduce the parameter and computational consumption of the plain Convolutional model \cite{resnet,vgg}. A series of subsequent works further improve the efficiency through depth-wise dilated Convolutions \cite{espnetv2}, network pruning \cite{prune}, neural structure search \cite{mobilenetv3}, channel shuffling \cite{shufflenet, shufflenetv2} and structure reparameterization \cite{fastvit, mobileone, repvit}. However, constrained by a limited receptive field, these efficient Convolutional networks lack global interactions between features, which degrades their performance. Although some works propose to increase the receptive field by increasing the size of the Convolutional kernel \cite{slak, replk}, however, it imposes high memory access cost and computational cost, and significantly slows down the inference speed. Considering that Transformers are capable of achieving global attention but introduce quadratic complexity with respect to the resolution of the input image, a series of recent works have attempted to combine the advantages of CNNs and Transformers to develop more efficient models \cite{efficientformer,swiftformer,fastvit}. For example, and EfficientFormer only adds self-attention at more deep stages to capture global context, SwiftFormer adds effective additive attention at each stage of the model. The success of this hybrid design motivates us to explore whether a more advanced backbone can be constructed by integrating Convolution and Mamba.

\begin{figure*}[t]
    \centering    \includegraphics[width=0.95\textwidth]{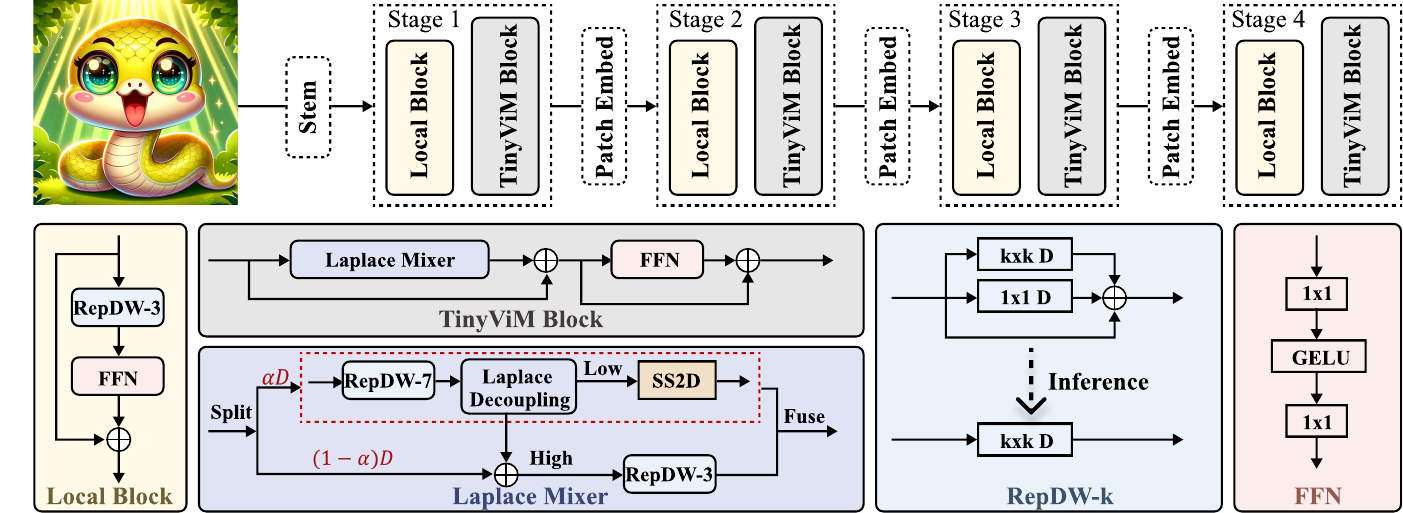}
    \caption{Overview of the proposed TinyViM, which has four stages and each stage consists of Local Blocks and TinyViM Blocks. The local block applies a reparameterized 3x3 Convolution to extract local features, and the TinyViM block is employed to capture the global context. The core component of the TinyViM block is the Laplace Mixer,  which decouples the frequencies of the features based on an efficient Laplace pyramid and passes the state only for the low-frequencies. The enhanced low-and high-frequency components are then integrated based on a frequency ramp Inception structure. Thus, the high- and low-frequency components of different stages are appropriately trade-off and the efficiency is further improved. SS2D denotes 2D selective scanning, which is used in VMamba \cite{vmamba}.}
    \label{fig:net}
\end{figure*}

\subsection{State Space Models}
As a typical architectural paradigm for sequence-to-sequence transformations, state-space models (SSMs) play a pivotal role in handling long-range dependencies in sequence data \cite{gu2021efficiently, gu2021combining, smith2022simplified,timepro}. Early SSMs have challenges in capturing long-term dependency relationships and a unified understanding of memory \cite{Graves2012}. To address these issues, Gu et al. enhance the ability of SSMs to capture extended dependencies using HiPPO initialization \cite{gu2020hippo} and solve the continuous-time memory problem based on a linear state space layer (LSSL) \cite{gu2021combining}. However, these SSMs are still limited by computational requirements.
Subsequent works propose a series of improvements including complex diagonal structures \cite{gu2022parameterization, gupta2022diagonal}, multiple-input multiple-output support \cite{smith2022simplified}, and diagonal-plus-low-rank arithmetic \cite{hasani2022liquid} to improve computational efficiency and generalization across tasks. In particular, these strategies have been extended to large representation models \cite{fu2022hungry, ma2022mega, mehta2022long}.
Recently Mamba \cite{mamba} introduce a selection mechanism that outperforms Transformers on large-scale real data through a data-dependent SSM layer with linearly scaling sequence lengths. 

The success of Mamba has caused the vision community to explore and apply it to various vision tasks including classification \cite{vim,vmamba}, detection \cite{quadmamba}, segmentation \cite{samba}, deraining \cite{fouriermamba} and multimodal learning \cite{vlmamba}. In particular, ViM \cite{vim} and VMamba \cite{vmamba} propose bi-directional scanning and cross-scanning strategies to cope with the non-sequential structure of visual data, respectively. QuadMamba \cite{quadmamba} proposes quad-tree scanning to capture local dependencies at different granularities. MSVMamba \cite{msvmamba} proposes multiscale two-dimensional scanning, which efficiently mitigates the long-range forgetting problem and reduces the computational cost. A highly related work is EfficientVMamba \cite{evmamba}, which improves the scanning efficiency of a model by an atrous-based selective scan approach. However, these works tend to show low throughput and fail to perform competitively with other advanced lightweight backbones based on Convolution or Transformers. This motivates us to design the TinyViM, which obtains satisfactory speed and performance based on Laplace frequency decoupling.

\section{Preliminaries}
\label{sec:pre}
\textbf{State Space Models (S4).} Inspired by the continuous system, State space models (SSMs) are proposed in deep learning as common sequence models \cite{gu2021efficiently}. These models transform an input sequence $x(t) \in \mathbb{R}^{L \times D}$ into an output sequence $y(t) \in \mathbb{R}^{L \times D}$ by utilizing a learnable hidden state $h(t) \in \mathbb{R}^{N}$. The process could be denoted as follows:
\begin{equation}
\label{eq1}
\begin{aligned}
    h'(t) &= \bm{A} h(t)+\bm{B} x(t),\\
    y(t) &= \bm{C}h(t),
\end{aligned} 
\end{equation}
where $\bm{A}\in \mathbb{R}^{N \times N}$ is the evolution parameter, $\bm{B}, \bm{C}\in \mathbb{R}^{N \times D}$ denote the learnable projection parameters, and $N$ is the state size. 

\noindent \textbf{Discretization.} The above continuous-time SSMs are not well compatible with deep learning algorithms. Therefore, discretization is needed to align the model with the sampling frequency of the input signal to improve computational efficiency \cite{gu2021combining}. Following the previous work \cite{gupta2022diagonal}, given the sampling time scale parameter $\bm{\Delta}$, the above continuous SSMs are discretized through zero-order hold rule, thus converting the continuous-time parameters ($\bm{A}$, $\bm{B}$) to their corresponding discrete counterparts ($\overline{\bm{A}}$, $\overline{\bm{B}}$):
\begin{equation}
\begin{aligned}
    \overline{\bm{A}} &= e^{\bm{\Delta}\bm{A}},\\
    \overline{\bm{B}} &= (\bm{\Delta}\bm{A})^{-1}(e^{\bm{\Delta}\bm{A}}-\bm{I})\cdot \bm{\Delta}\bm{B}.
\end{aligned} 
\end{equation}
Then, The discretized formulation of Eq.~\ref{eq1} is formulated as:
\begin{equation}
\label{eq3}
\begin{aligned}
    h_t &= \overline{\bm{A}} h_{t-1}+\overline{\bm{B}} x_t,\\
    y_t &= \bm{C}h_t,
\end{aligned} 
\end{equation}
where $\overline{\bm{A}}\in \mathbb{R}^{N \times N}$, $\overline{\bm{B}}\in \mathbb{R}^{N \times D}$. In order to improve the computational efficiency, the iterative process described in Eq.~\ref{eq3} can be performed by the parallel computing mode of global convolution \cite{mamba}:
\begin{equation}
\begin{aligned}
    y &= x \circledast \overline{\bm{K}},   \\
   \text{with} \quad  \overline{\bm{K}} &= (\bm{C}\overline{\bm{B}},\bm{C}\overline{\bm{A}\bm{B}},\cdots, \bm{C}\overline{\bm{A}}^{L-1}\overline{\bm{B}}),
\end{aligned}
\end{equation}
where $\circledast$ denotes the convolution operation, and $\overline{\bm{K}}\in \mathbb{R}^{L} $ serves as the kernel of the SSMs.

\noindent \textbf{Selective State Space Models (S6).} Conventional SSMs (i.e., S4) have been implemented to capture sequence context under linear time complexity, despite the fact that they are constrained by static parameterization and cannot perform content-based reasoning. 
To address this problem, the selective state space model (i.e., Mamba \cite{mamba}) has been proposed, which allows the model to selectively propagate or forget information based on the latitude of the current token along the length of the sequence by simply setting the parameters of the SSM as a function of the inputs.
 In S6, the parameters $\bm{B}$, $\bm{C}$, and $\Delta$ are computed directly from the input sequence $x(t)$, thus enabling sequence-aware parameterization.

\section{Method}
\subsection{Analysis of Vanilla Mamba Block}
\label{sec:analysis}
As analyzed in Sec.~\ref{sec:pre}, Mamba is able to model global context information based on linear complexity with respect to sequence length through pixel-by-pixel state transfer, which shows potential in lightweight model design. Moreover, considering the mobile-friendly and localized feature extraction capability of Convolution, a natural intuition arises: can we build a lightweight and efficient visual backbone based on Convolution and Mamba blocks? Although there have been some works \cite{evmamba} that attempt to do so, the performance and efficiency shown are not competitive with Convolution \cite{mobileone} or Transformer-based models \cite{fastvit,swiftformer} so far.

\noindent \textbf{Baseline Construction.} In response to this intuition, we first build a baseline architecture using Convolution (the local block in Fig.~\ref{fig:net}) and a vanilla Mamba block (applying SS2D in VMamba \cite{vmamba} to adapt to the visual data), i.e., the Baseline in Table~\ref{tab:ab}. 
Note that the baseline structure is identical to the final structure of TinyViM-S, which is designed with reference to \cite{swiftformer}, and specific structural configurations can be found in the Supplementary Material. The only difference between Baseline and TinyViM-S is that we replace the TinyViM blocks with SS2D-based vanilla Mamba blocks.
We are surprised to find that this version achieves a Top-1 accuracy of 79.1\%, which is significantly better than the purely Convolutional (i.e., Conv only) version. This can be explained by the fact that Mamba can improve the expressive ability of the model by capturing the global context at each stage. However, the low throughput limits the use of the model in real-time applications. We then attempt to improve the baseline based on qualitative and quantitative analysis.

\begin{table}[t]
  \centering
  \caption{
    Effect of frequency input Mamba block. Throughput is tested on a Nvidia V100 GPU with maximum power-of-two batch size that fits in memory. The Top-1 accuracy is the average of three experiments to exclude random fluctuations.
  }
  \small
  \begin{tabular}{c|c|c|c}
  \toprule
   Variants &GMACs & Throughput & Top-1 \\
   \hline
   Baseline &0.96 &1673 &79.1\\
   Conv only &0.88 &3212 &77.5\\
   \rowcolor[gray]{0.92}
   \textbf{Low only}& \bf0.93 & \bf2574 & \bf79.0\\
   High only &0.96 &1377 &78.6 \\
   Low + High &0.97 &1509 &79.1\\
  \bottomrule
  \end{tabular}
  \vspace{-3mm}
  \label{tab:ab}
\end{table}

\noindent \textbf{Qualitative Analysis.} We first perform a spectral analysis of the feature maps before and after inputting the Mamba block in the baseline, as shown in Fig.~\ref{fig:freq}. It shows that the Mamba block has strong low-frequency capture ability under the Convolution-Mamba hybrid architecture, but loses high-frequency information, such as local edges and textures. This low-frequency preference impairs the performance and efficiency of vision Mamba because 1) Populating each stage with low-frequency information degrades the high-frequency component, weakening the fine-grained recognition ability of visual Mamba, and 2) Additional high-frequency component input to a Mamba block impairs the efficiency and parallel computation of the model. A feasible strategy for this phenomenon is that we decouple the high and low frequencies of the features and input only the low-frequency components of the features into the Mamba block for hidden state transfer.
Therefore, we decouple the high and low frequencies of the feature maps based on the Laplace pyramid, which is a more efficient architecture compared to other methods such as the cascaded wavelet transform, to verify the effectiveness of our strategy.

\noindent \textbf{Quantitative Analysis.} We next set the inputs of the Mamba block for each stage to be low-frequency only, high-frequency only, high and low-frequency in parallel to explore the effect of the corresponding variants on the classification accuracy, respectively. As shown in Table~\ref{tab:ab}, when only low frequencies are used as inputs, the model achieves 79.0\% accuracy, reduces the computational consumption of the model and achieves 1.5 times higher throughput. This verifies that we can significantly improve the efficiency of the model without compromising the classification accuracy by using only low-frequency inputs to Mamba. Therefore, based on the above qualitative and quantitative analyses, we design the Laplace mixer to optimize Mamba in the construction of a lightweight vision backbone.

\setlength{\tabcolsep}{3pt}
\begin{table*}[t]
	\begin{minipage}{0.68\linewidth}
		\vspace{0pt}
		\caption{{\textbf{Classification performance on ImageNet-1K.} Following~\cite{efficientvit}, throughput is tested on Nvidia V100 GPU with maximum power-of-two batch size that fits in memory.}}
		\label{tab:comparison}
		\centering
		\scriptsize
		\begin{tabular}{ccccccccccc}
			\toprule
			\multirow{2}{*}{Model}    & \multirow{2}{*}{Conference}       & \multirow{2}{*}{Type}      & \multirow{2}{*}{\makecell{Params $\downarrow$ \\ (M)}} & \multirow{2}{*}{GMACs $\downarrow$}  & \multirow{2}{*}{\makecell{Throughput $\uparrow$ \\ (im/s)}} & \multirow{2}{*}{Reso.} & \multirow{2}{*}{\makecell{Top-1 $\uparrow$ \\ (\%)}}  \\
			\\
			\hline
			\hline
			MobileOne-S1 \cite{mobileone} &\textcolor{blue}{CVPR'23}& CNN &4.8 & 0.8 &3545 &224&75.9 \\
			SwiftFormer-S~\cite{swiftformer}&\textcolor{blue}{ICCV'23} & CNN+ViT & 6.1 & 1.0 &2626  & 224 & 78.5 \\
			EfficientFormerV2-S0~\cite{efficientformerv2}  &\textcolor{blue}{ICCV'23}&  CNN+ViT   & 3.5      &  0.4          &985  & 300 & 75.7  \\
			EfficientVMamba-T \cite{evmamba} &\textcolor{blue}{Arxiv'24}&Mamba&6.1 &1.0&1396 &224 &76.5 \\
			Vim-Ti \cite{vim}&\textcolor{blue}{ICML'24}& Mamba &7.0 & 1.5 & 617&300 &76.1\\
			QuadMamba-Li \cite{quadmamba}& \textcolor{blue}{NeurIPS'24} &Mamba &5.4 &0.8 &1190 &224 &74.2\\
			MSVMamba-N \cite{msvmamba}&\textcolor{blue}{NeurIPS'24}&Mamba &7.0 &0.9 &1283 &224 &77.3 \\
			MobileMamba-S6 \cite{mobilemamba} &\textcolor{blue}{CVPR'25} &CNN+Mamba &15.0 &0.7 &3650 &224 &78.0 \\
			\rowcolor[gray]{0.92}
			\textbf{TinyViM-S} &- &\textbf{CNN+Mamba}  &\textbf{5.6} &\textbf{0.9} &\textbf{2563} &\textbf{224} &\bf 79.2\\ %
			\hline
			PoolFormer-S12 \cite{metaformer}&\textcolor{blue}{CVPR'22} &Pool &12.0&2.0&1902 &224&77.2 \\
			MobileOne-S3 \cite{mobileone}&\textcolor{blue}{CVPR'23} & CNN &10.1 &1.9 &1900 &224 &78.1 \\
			MobileOne-S4 \cite{mobileone}&\textcolor{blue}{CVPR'23} & CNN &14.8 &3.0 &1223 &224 &79.4 \\
			Agent-PVT-T \cite{agent} & \textcolor{blue}{ECCV'24} &ViT &11.6 &2.0 &1447 &224&78.4 \\
			EfficientVMamba-S \cite{evmamba}&\textcolor{blue}{Arxiv'24} &Mamba &11.0 &1.3 &674 &224 &78.7 \\
			QuadMamba-T \cite{quadmamba}& \textcolor{blue}{NeurIPS'24} &Mamba &10.0 &2.0 &585 &224 &78.2\\
			MSVMamba-M \cite{msvmamba}&\textcolor{blue}{NeurIPS'24} &Mamba &12.0 &1.5 &957 &224 &79.8 \\
			MobileMamba-B1 \cite{mobilemamba} &\textcolor{blue}{CVPR'25} &CNN+Mamba &17.1 &1.1 &2097 &256 &79.9 \\
			\rowcolor[gray]{0.92}
			\textbf{TinyViM-B} &- &\textbf{CNN+Mamba}  &\bf11.0 &\bf1.5 &\bf1851 &\bf224 &\bf 81.2\\ %
			\hline
			PoolFormer-S36 \cite{metaformer}&\textcolor{blue}{CVPR'22} &Pool &31.0 &5.2 &667 &224&81.4\\
			InceptionNext-T \cite{inceptionnext}&\textcolor{blue}{CVPR'24} &CNN & 28.0 &4.2 &987 &224 &82.3 \\
			Agent-PVT-S \cite{agent} & \textcolor{blue}{ECCV'24} &ViT &20.6 &4.0 &686 &224 &82.4 \\
			FastViT-SA24 \cite{fastvit}&\textcolor{blue}{ICCV'23} &CNN+ViT &20.6 &3.8 &861 &224 &82.6 \\
			EfficientVMamba-B \cite{evmamba} &\textcolor{blue}{Arxiv'24} & Mamba &33.0 &4.0 &580 &224 &81.8 \\
			LocalVmamba-T \cite{localmamba}&\textcolor{blue}{Arxiv'24} &Mamba &26.0 &5.7 &- &224 &82.7 \\
			ViM-S \cite{vim} &\textcolor{blue}{ICML'24} &Mamba &26.0 &5.1 &201 &224 &80.5 \\
			VMamba-T \cite{vmamba} &\textcolor{blue}{NeurIPS'24} &Mamba &30.0 &4.9 &383 &224&82.6 \\
			QuadMamba-S \cite{quadmamba}& \textcolor{blue}{NeurIPS'24} &Mamba &31.0 &5.5 &541 &224 &82.4\\
			MobileMamba-B4 \cite{mobilemamba} &\textcolor{blue}{CVPR'25} &CNN+Mamba &17.1 &4.3 &513 &512 &82.5 \\
			\rowcolor[gray]{0.92}
			\textbf{TinyViM-L} &- &\textbf{CNN+Mamba}  &\bf31.7 &\bf4.7 &\bf843 &\bf224 &\bf 83.3\\
			\bottomrule
		\end{tabular}
	\end{minipage}
	\hspace{0.02\linewidth}
	\begin{minipage}{0.3\linewidth}
		\caption{
			Ablation of frequency ramp inception design.
		}
		\vspace{-6pt}
		\label{tab:ramp}
		\scriptsize
		\centering
		\begin{tabular}{cc|c|c|c}
			\toprule
			Inception& Ramp &GMACs & Throughput & Top-1 \\
			\hline
			\XSolidBrush &\XSolidBrush &0.94 &2478 &79.2\\
			\CheckmarkBold &\XSolidBrush &0.93 &2574 &79.0\\
			\rowcolor[gray]{0.92}
			\CheckmarkBold & \CheckmarkBold& \bf0.93 &\bf2563 &\bf79.2 \\
			\bottomrule
		\end{tabular}
		\vspace{3pt}
		\caption{
			Ablation of the partition coefficient $\alpha$ of four stages.
		}
		\vspace{-6pt}
		\label{tab:alpha}
		\begin{tabular}{cccc|c|c|c}
			\toprule
			$\alpha_1$& $\alpha_2$& $\alpha_3$& $\alpha_4$ &GMACs & Throu. & Top-1 \\
			\hline
			0.75&0.5 &0.5 &0.25 &0.94 &2591 &78.8\\
			0.5 &0.5 &0.5 &0.5 &0.93 &2574 &79.0\\
			\rowcolor[gray]{0.92}
			\bf0.25&\bf0.5 &\bf 0.5 &\bf 0.75 &\bf0.93 &\bf 2563 &\bf79.2 \\
			\bottomrule
		\end{tabular}
		\vspace{3pt}
		\caption{
			Ablation of the convolution kernel size in Laplace Mixer.
		}
		\centering
		\vspace{-6pt}
		\label{tab:kernel}
		\begin{tabular}{cc|c|c|c}
			\toprule
			kernel & axial&GMACs & Throughput & Top-1 \\
			\hline
			3 &\XSolidBrush &0.94 &2510 &79.0\\
			5 &\CheckmarkBold &0.93 &2598 &79.0\\
			\rowcolor[gray]{0.92}
			\bf7 &\CheckmarkBold &\bf0.93 &\bf2563 &\bf79.2\\
			9 &\CheckmarkBold &0.94 &2479 &79.2\\
			\bottomrule
		\end{tabular}
		\vspace{3pt}
		\caption{
			Ablation of re-parameterization.
		}
		\vspace{-6pt}
		\label{tab:rep}
		\begin{tabular}{cc|c|c|c}
			\toprule
			Variant & Rep&GMACs & Throu. & Top-1 \\
			\hline
			\rowcolor[gray]{0.92}
			\textbf{TinyViM-S} & \CheckmarkBold &\bf0.93 &\bf2563 &79.2\\
			TinyViM-S &\XSolidBrush&0.93 &2563 &79.2\\
			\hline
			\rowcolor[gray]{0.92}
			\textbf{TinyViM-L} & \CheckmarkBold &\bf4.72 &\bf843 &\bf83.3\\
			TinyViM-L &\XSolidBrush &4.72 &843 &83.1\\
			\bottomrule
		\end{tabular}
	\end{minipage}
	\vspace{-3mm}
\end{table*}

\subsection{Laplace Mixer}
Previous work such as EfficientVMamba \cite{evmamba} uses dilated-based selective scanning to improve the efficiency of vanilla Mamba \cite{mamba}. Despite reducing the number of input tokens, EfficientVMamba performs unsatisfactorily on various visual tasks \cite{evmamba}. We argue that the potential of Mamba in lightweight backbones is not fully exploited simply by modifying the scanning path in the image domain. Based on the analysis in Sec.~\ref{sec:analysis}, we reconstruct the Mamba block based on the frequency decoupling and propose an elegant Laplace mixer.
Furthermore, consider that in a general visual backbone, the bottom layer is more concerned with capturing high-frequency details, while the top layer is more concerned with modeling low-frequency global information \cite{resnet}. As in humans, by capturing details in high-frequency components, the lower layers can capture the basic visual features and gradually collect local information to achieve a global understanding of the input. In particular, feature redundancy has been shown to exist in existing networks \cite{ghostnet}, and separate low-frequency and high-frequency processing for all channels is unnecessary. Therefore, we design a frequency ramp inception structure which reduces the channel size of the high-frequency branches as the depth of the network increases.

Specifically, for the input feature $X \in \mathbb{R}^{H \times W \times D}$, where H, W and D denote the height, width and channels of $X$, we first split along the channel dimension and divide it into a low-frequency input $X_l \in \mathbb{R}^{H \times W \times {\alpha}D}$ and a high-frequency input $X_h  \in \mathbb{R}^{H \times W \times (1-\alpha)D}$. $\alpha$ denotes the partition coefficient, which determines the number of channels in the low-frequency branch. Then, for the low-frequency branch, we adopt a simple Laplace pyramid architecture. Specifically, we average pool $X_l$ to obtain the corresponding low-frequency components $X_{ll}  \in \mathbb{R}^{\hat{H} \times \hat{W} \times {\alpha}D}$ and shrink the number of input tokens. Next, we recover the resolution of $X_{ll}$ based on the nearest neighbor upsampling and perform pixel-by-pixel subtraction from $X_l$ to obtain the high-frequency component of $X_l$, i.e., $X_{lh}$. This process can be formulated as,
\begin{equation}
    X_{ll} = \mathrm{Pool}(X_l), \quad X_{lh} = X_l - \mathrm{Upsample}(X_{ll}).
\end{equation}
Then, we concatenate $X_{lh}$ and $X_h$ along the channel to obtain the high-frequency input $X_{hh} \in \mathbb{R}^{H \times W \times D}$, and apply a small depth-wise Convolution kernel, which usually tends to respond to the high frequencies in the input \cite{conv1, conv2}, enhancing the high-frequency component of $X_{hh}$. As for the low-frequency input $X_{ll}$, we input it into the SS2D block for cross-scanning to capture the global context. The process of frequency separation processing can be described as,
\begin{equation}
    \hat{X}_{ll} = \mathrm{SS2D}(X_{ll}), \quad \hat{X}_{hh} = \mathrm{Rep_3}(X_{hh}).
\end{equation}
where $\mathrm{Rep_3}$ denotes the reparameterized $3 \times 3$ convolution.
Finally, we sum up $\hat{X}_{ll}$ and $\hat{X}_{hh}$ of the corresponding channels at the element level, followed by a $1 \times 1$ convolution for fusion. Through the frequency decoupling process and frequency ramp inception, the Laplace mixer can effectively weigh and integrate the high and low-frequency components of all layers. As a result, TinyViM achieves a better balance between accuracy and efficiency.

\subsection{Overall Architecture}
TinyViM employs a similar multi-scale backbone design similar to previous works \cite{efficientformer, swiftformer}. As shown in Fig.~\ref{fig:net}, given an image $I \in \mathbb{R}^{H \times W \times 3}$, a simple stem is employed to output $X_1 \in \mathbb{R}^{\frac{H}{4}\times \frac{W}{4} \times D_1}$ feature map. $H$, $W$ denote the height and width of the input image respectively, and $D_1$ is the the channel number of $X_1$. Note that the stem is implemented with two $3 \times 3$ convolutions with a stride of 2. Then, the output feature maps are fed into the first stage, which begins with Local Blocks. The local block consists of a reparameterized $3 \times 3$ convolution and a feed-forward network (FFN, implemented by two consecutive $1 \times 1$ convolutions) that achieves local feature extraction and feature channel mixing, respectively. 
The local block can be formulated as,
\begin{equation}
    \hat{X}_1 = \mathrm{Conv_1}(\mathrm{GeLU}(\mathrm{Conv_1}(\mathrm{Rep_3}(X_1)))),
\end{equation}
where $\mathrm{Conv_1}$ and $\mathrm{Rep_3}$ denote the $1 \times 1$ and reparameterized $3 \times 3$ convolution, respectively.
The features are then fed into the TinyViM block for global context-awareness. Similarly, the TinyViM block consists of a Laplace mixer and FFN for global feature extraction and channel mixing, respectively. The TinyViM Block can be described as,
\begin{equation}
    \hat{X'}_1 = \mathrm{LaplaceMixer}(\hat{X}_1)+\hat{X}_1,
\end{equation}
\begin{equation}
    \tilde{X}_1 = \mathrm{Conv_1}(\mathrm{GeLU}(\mathrm{Conv_1}(\hat{X'}_1)))+\hat{X'}_1.
\end{equation}
There is a Patch Embedding layer between two consecutive stages for increasing the channel dimension and reducing the resolution of feature maps. Next, the resulting feature maps are subsequently fed into the second, third, and fourth stages, producing $\frac{H}{8}\times \frac{W}{8} \times D_2$, $\frac{H}{16}\times \frac{W}{16} \times D_3$ and $\frac{H}{32}\times \frac{W}{32} \times D_4$ feature maps, respectively.

\section{Experiments}
\label{sec:exp}
We evaluate the proposed TinyViM across four down-stream tasks: classification on ImageNet-1K dataset \cite{imagenet}, object detection and instance segmentation on MS-COCO 2017 dataset \cite{coco}, and semantic segmentation on ADE20K dataset \cite{ade}. All experiments are implemented following the common training settings in \cite{swiftformer}. Due to page limits, we provide a description of the dataset, the experimental setup, details of the model's architecture and more qualitative analysis in the Supplementary Material.
\subsection{Image Classification}
\begin{figure}[t]
	\vspace{-2mm}
	\centering    \includegraphics[width=0.45\textwidth]{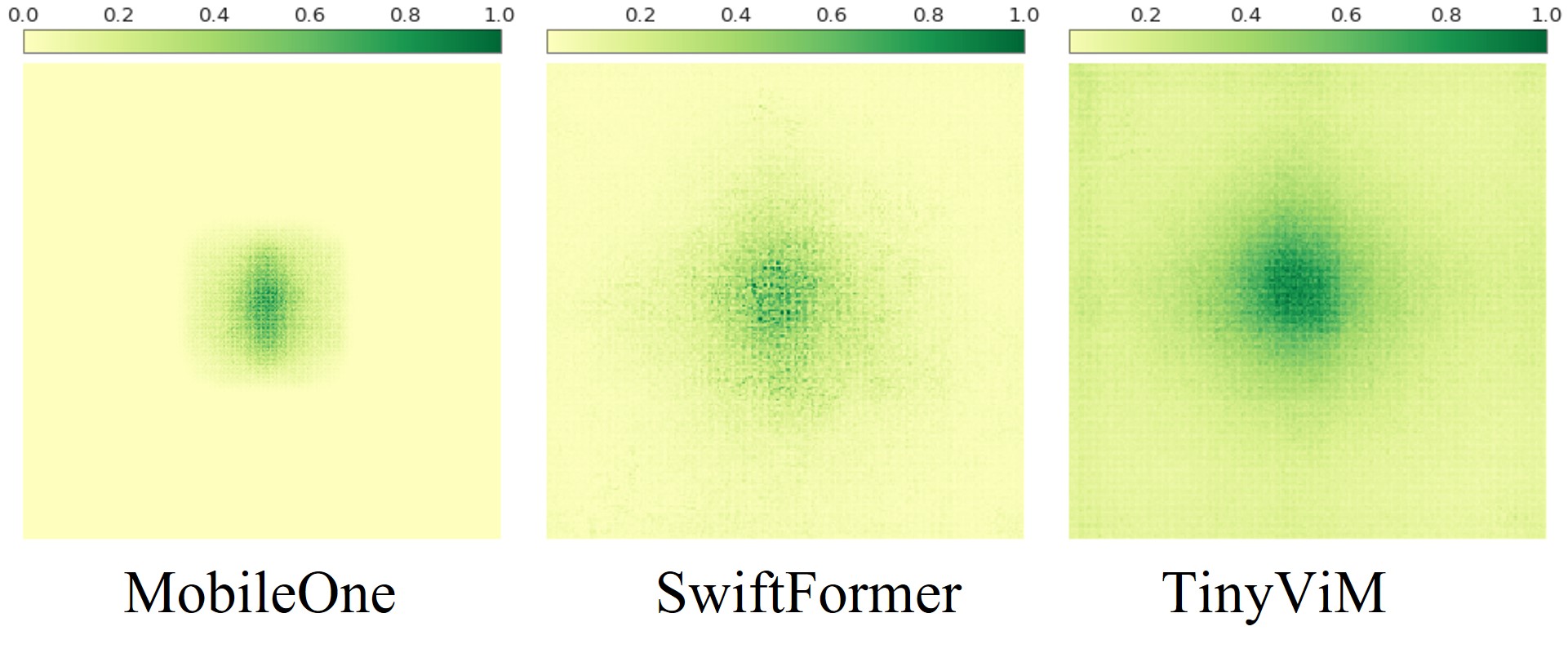}
	\caption{The ERF of MobileOne, SwiftFormer and TinyViM. Our TinyViM effectively obtains large ERFs with Laplace Mixer.}
	\label{fig:erf}
		\vspace{-1mm}
\end{figure}

We evaluate TinyViM's performance in image classification on ImageNet-1K \cite{imagenet} with some advanced Convolution, ViT and Mamba-based methods, and the comparison results are summarized in Table~\ref{tab:comparison}. With similar GMACs and throughput, TinyViM-S achieves a top-1 accuracy of 79.2\%, outperforming SwiftFormer-S by 0.7\%. In particular, TinyViM retains its performance advantages even at small and large scales. For example, TinyViM-B achieves an accuracy of 81.2, which outperforms the convolution-based model MobileOne-S4 by 1.8\%, the ViT-based model Agent-PVT-T by 2.8\%,  and the Mamba-based model MSVMamba-M by 1.4\%, respectively. These comparison results show the excellent performance of TinyViM. In addition, we compare the visualisation of  Effective Receptive Field (ERF) \cite{erf} with the classical lightweight backbone, as shown in Fig. \ref{fig:erf}. The comparison models include the pure Convolution-based MobileOne-S3 and the Convolution-ViT hybrid model SwiftFormer-S. It can be observed that TinyViM-S possesses a larger ERF at a similar size. This suggests that TinyViM is able to effectively model the global context through the Laplace Mixer, thus improving the semantic recognition ability of the model.

In terms of computational efficiency, TinyViM-S achieves a throughput of 2574 images/s, which is significantly better than other visual Mamba models. For example, the throughput of EfficientVmamba is only 1396 images/s, which is only about half of the throughput of TinyViM. This huge efficiency advantage persists on TinyViM-B and TinyViM-L, where the throughput is 2.7 and 1.5 times higher than that of the similarly scaled EfficientVmamba, respectively. Compared with other convolutional and ViT-based models, TinyViM maintains its lead in performance while its throughput is highly competitive. It should be emphasized that this competitiveness is significantly improved on A100 GPUs. This is due to the fact that Mamba has a number of hardware-friendly improvements on the A100 such as triton, which are not yet supported on the V100 GPUs. In summary, the results show that TinyViM strikes a better balance between performance and efficiency than other state-of-the-art lightweight backbones, and offers high value in real-time applications.

\subsection{Object Detection and Instance Segmentation}
We evaluate the proposed TinyViM for object detection and instance segmentation using the Mask R-CNN \cite{maskrcnn} framework. As shown in Table~\ref{tab:coco}, TinyViM outperforms competitive models such as the pooling-based model PoolFormer, the Transformer-based models FastVit and SwiftFormer, and the Mamba-based model EfficientVMamba in both $\text{AP}^{box}$ and $\text{AP}^{mask}$ metrics. Specifically, compared to recent models SwiftFormer-L1, TinyViM has 1.1\% and 0.6\% improvement on $\text{AP}^{box}$ and $\text{AP}^{mask}$, respectively, and has a larger throughput. Compared to EfficientVMamba-S, TinyViM-B has more significant advantages, i.e., 3.0 and 2.0 improvement in  $\text{AP}^{box}$ and $\text{AP}^{mask}$, respectively, with a 1.7 times throughput. When comparing with larger models such as FastViT-SA24 and EfficientVMamba-b, TinyViM still has an advantage in throughput, and shows 2.5\% and 0.8\% improvements in $\text{AP}^{box}$, respectively. These results demonstrate the superiority of TinyViM when transferring to detection and instance segmentation tasks.

\begin{table*}[t]
	\centering
	\caption{
		\textbf{Object detection \&
			instance segmentation} results on MS COCO 2017 with the Mask RCNN framework.
		\textbf{Semantic segmentation} results on ADE20K by integrating models into Semantic FPN. 
	}
	\vspace{-2mm}
	\scriptsize
	\resizebox{0.9\linewidth}{!}{
		\begin{tabular}{c|c|ccc|ccc|c}
			\toprule
			\multirow{2}{*}{Backbone} & \multirow{2}{*}{\makecell{Throughput $\uparrow$ \\ (img/s)}} & \multicolumn{3}{c|}{Object Detection} & \multicolumn{3}{c|}{Instance Segmentation} & \multicolumn{1}{c}{Semantic}  \\ \cline{3-9}
			&   & AP$^{box}$    & AP$^{box}_{50}$   & AP$^{box}_{75}$   & AP$^{mask}$    & AP$^{mask}_{50}$   & AP$^{mask}_{75}$   & mIoU   \\
			\hline
			\hline
			EfficientVMamba-T \cite{evmamba}&168 &35.6 &57.7 &38.0 &33.2 &54.4 &35.1 &- \\
			\rowcolor[gray]{0.92}
			\textbf{TinyViM-S} &\bf262 &\bf38.7 &\bf61.1 &\bf41.3 &\bf35.7 &\bf57.1 &\bf37.9 &\bf38.9  \\
			\hline
			
			PoolFormer-S12~\cite{metaformer}        & 171       & 37.3   & 59.0    & 40.1    & 34.6   & 55.8    & 36.9     & 37.2  \\
			EfficientFormer-L1~\cite{efficientformer}   & 197 & 37.9   & 60.3    & 41.0    & 35.4   & 57.3    & 37.3   &  38.9  \\
			FastViT-SA12 \cite{fastvit} &162 &38.9 &60.5 &42.2 &35.9 &57.6 &38.1 &38.0 \\
			SwiftFormer-L1 \cite{swiftformer}&174&41.2&63.2 &44.8 &38.1 &60.2 &40.7 &41.1 \\
			EfficientVMamba-S \cite{evmamba}&104 &39.3 &61.8 &42.6 &36.7 &58.9 &39.2 &- \\
			MobileMamba-B1 \cite{mobilemamba} &- &40.6 &61.8 &43.8 &37.4 &58.9 &39.9 & 40.7 \\
			\rowcolor[gray]{0.92}
			\textbf{TinyViM-B} &\bf180 &\bf42.3 &\bf64.2 &\bf46.3 &\bf38.7 &\bf61.1 &\bf41.3 &\bf41.9  \\
			\hline
			PoolFormer-S36~\cite{metaformer}      &  87      & 41.0   & 63.1    & 44.8    & 37.7   & 60.1    & 40.0  &  42.0 \\
			EfficientFormer-L3~\cite{efficientformer} &117     & 41.4   & 63.9    & 44.7    & 38.1   & 61.0    & 40.4   &  43.5 \\
			FastViT-SA24 \cite{fastvit} &110& 42.0 &63.5 &45.8 &38.0 &60.5 &40.5&41.0 \\
			SwiftFormer-L3 \cite{swiftformer} &120&42.7 &64.4 &46.7 &39.1 &61.7 &41.8 &43.9 \\
			EfficientVMamba-B \cite{evmamba}& 90 &43.7 &66.2 &47.9 &40.2 &63.3 &42.9 &- \\
			MobileMamba-B4 \cite{mobilemamba} &- &40.1 &61.8 &43.0 &36.9 &58.6 &39.2 & 42.5 \\
			\rowcolor[gray]{0.92}
			\textbf{TinyViM-L} &\textbf{111} &\textbf{44.5} &\textbf{66.4} &\bf48.6 &\bf40.7 &\bf63.6 &\bf43.8 &\bf44.2\\
			\bottomrule
		\end{tabular}
	}
	\label{tab:coco}
\end{table*}

\subsection{Semantic Segmentation}
Table~\ref{tab:coco} shows the semantic segmentation results of TinyViM as a backbone in comparison with other state-of-the-art backbones, where Semantic FPN \cite{fpn} serves as the decode head. Specifically, compared with recent methods such as SwiftFormer, TinyVim obtains 0.8\% and 0.3\% improvement in base and large scale, respectively. Compared with the method FasterViT, TinyViM has more significant advantages, with 3.9\% and 3.2\% improvements, respectively. The experimental results verify the effectiveness of TinyViM in semantic segmentation tasks.

\subsection{Ablation Studies}

\begin{figure}[t]
	\vspace{-2mm}
	\centering    \includegraphics[width=0.48\textwidth]{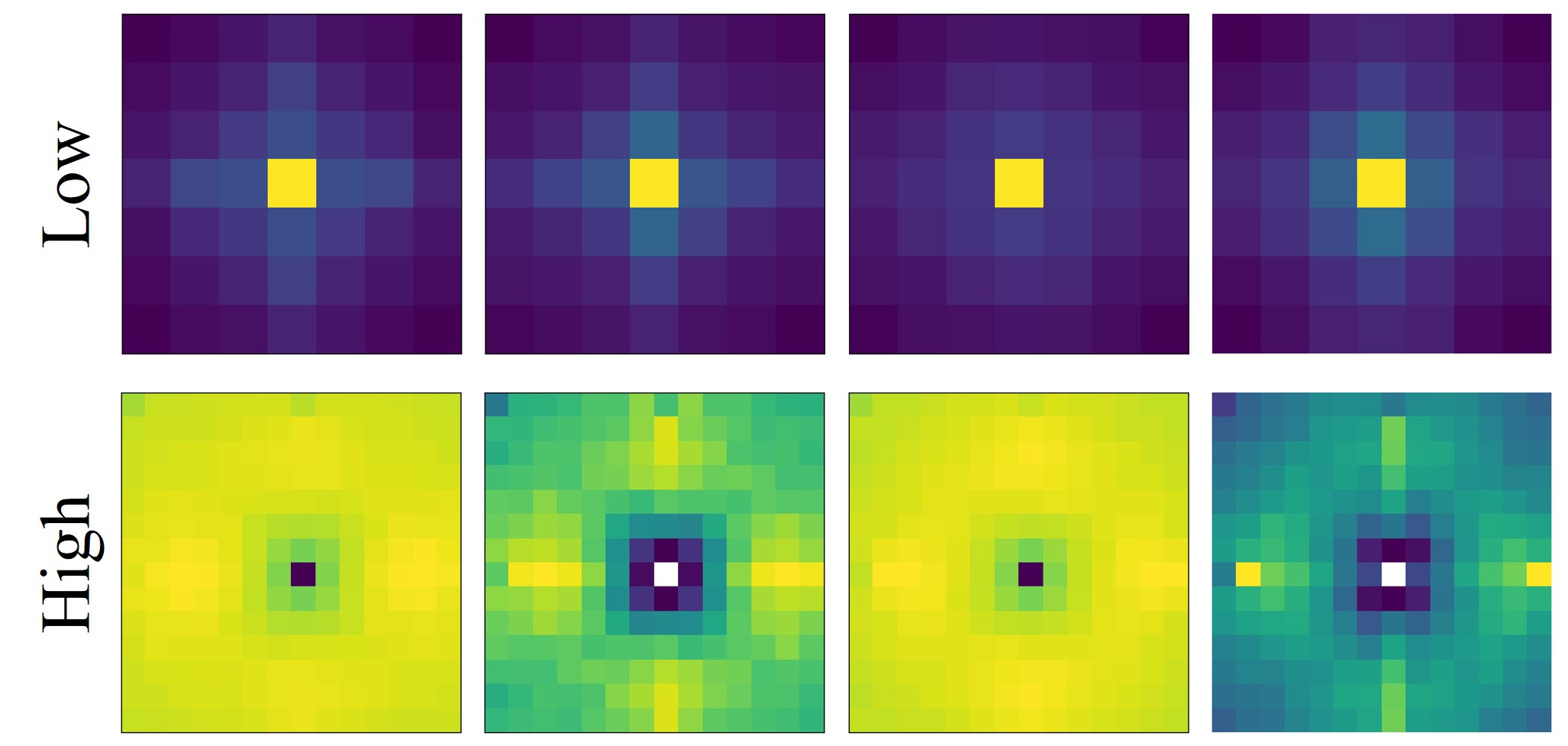}
	\caption{Fourier spectrum of TinyViM for the \textbf{Low} frequency branch and \textbf{High} frequency branch. It can be observed that we achieve the decoupling of the high and low frequency components of the features and specifically enhance them separately.}
	\label{fig:res}
	\vspace{-2mm}
\end{figure}

\noindent \textbf{Effect of the Frequency Decoupling.} 
In order to investigate the effectiveness of the frequency decoupling strategy, we perform spectral visualization of the feature maps of the low-frequency and high-frequency branches, respectively. As shown in Fig.~\ref{fig:res}, by applying Mamba, the feature maps have higher concentrations at low frequencies; by applying depth-wise convolution, the high frequencies of the model are effectively preserved. This proves that the Laplace mixer can effectively decouple the high and low frequency components of the features and specially enhance them respectively.

\noindent \textbf{Ablation of the Frequency Ramp Inception.}
We explore the effectiveness of the frequency ramp inception design, as shown in Table~\ref{tab:ramp}. We set up three experimental variants: variant 1, we implement Laplace frequency decoupling for all channels of the feature map, and input low and high frequencies into Mamba and depth-wise convolution, respectively; variant 2, we adopt a non-ramp structure, i.e., we set the segmentation coefficient $\alpha$ of the four stages to be 1/2; and variant 3, we gradually increase $\alpha$ with the increase of the stage of the model. In the implementation of this paper, the $\alpha$ of the four stages are 0.25, 0.5, 0.5 and 0.75, respectively. It can be observed that, even in the small size, the throughput of variant 1 is significantly reduced and more GMACs are required. Variant 2, although more efficient, has a decreased accuracy due to the lack of adjusting the weights of the low-frequency and high-frequency components in different stages. We also explore the inverse ramp structure, as shown in Table \ref{tab:alpha}. The frequency ramp used in variant 3 effectively weighs and integrates the high and low frequency information, thus helping the model to achieve a better balance between performance and efficiency. 

\noindent \textbf{Ablation of the convolution kernel size in Laplace Mixer.} ViM and VMamba use a $3 \times 3$ depthwise convolution before scanning. However, a larger convolution kernel is beneficial for parameters (i.e., $\bm{B}$, $\bm{C}$) generation due to the large receptive fields, although it impair the efficiency of the model. Therefore, we introduce axial depthwise convolution, which increases the convolution's receptive field along the scan path without decreasing the model efficiency. We also explore $3 \times 3$ depthwise convolution and axial convolution with kernel sizes 5 and 9, as shown in Table \ref{tab:kernel}. The model achieves the best balance between performance and efficiency when the kernel size is 7.

\noindent \textbf{Ablation of re-parameterization.} As shown in Table \ref{tab:rep}, the reparameterization has no improvement on TinyViM-S, but brings an increase of 0.2 to TinyViM-L. Considering that the reparameterization technique does not harm the inference efficiency of the model, we apply it for TinyViM.

\section{Conclusion}
Motivated by Mamba's ability to achieve global context modeling based on linear complexity, we construct a tiny and efficient hybrid vision Mamba TinyViM by combining the mobile-friendly convolution and Mamba in this paper. In contrast to previous vision Mamba efforts that simply modify scan paths in the image domain, whose performance could not compete with Convolution or Transformer based models with similar scale, TinyViM attempts to build a lightweight backbone from a frequency decoupling perspective. Specifically, we verify that the Mamba block mainly models low-frequency information in a Convolution-Mamba hybrid architecture after spectral analysis and quantitative experiments. To this end, we design a frequency decoupling strategy to preserve the high-frequency information and improve the low-frequency modeling efficiency of Mamba. In particular, we introduce a frequency ramp inception to weigh the different strengths of the low-frequency and high-frequency components at different stages of the model, so as to help the model achieve a better balance between accuracy and efficiency. Extensive experiments have shown that TinyViM outperforms other Convolution, Transformer and Mamba-based models. 
In future work, we plan to apply it as a lightweight backbone to replace the image encoder in SAM to explore the potential of vision Mamba in multi-modal scenarios.
 
{
    \small
    \bibliographystyle{ieeenat_fullname}
    \bibliography{main}
}
\clearpage
\setcounter{page}{1}
\maketitlesupplementary

\section{Dataset and Implementation Details}
\noindent \textbf{ImageNet} \cite{imagenet} is a large image categorization dataset for computer vision research, containing about 1.2 million labeled images of about 1000 categories. All of our models are trained on the Imagenet-1k dataset for 300 epochs using the AdamW optimizer and an initial learning rate of $2e^{-3}$. We use an image resolution of 224x224 for both training and testing. In addition, we use the same teacher model (i.e., the RegNetY-16GF model with a top-1 accuracy of 82.9\%) as in \cite{efficientformer,swiftformer} for distillation. All experiments are conducted on 8 NVIDIA V100 GPUs and the throughput is tested with maximum power-
of-two batch size that fits in memory.

\noindent \textbf{MS-COCO 2017} \cite{coco} is a large-scale real-world image and annotation dataset that contains 118K training and 5K validation images with 80 object classes. We use TinyViM as the backbone for feature extraction and apply the Mask-RCNN \cite{maskrcnn} framework for object detection and instance segmentation. Similar to \cite{efficientformer},  we finetune our TinyViM for 12 epochs with an image resolution of 1333 $\times$ 800 and batch
size of 32. The AdamW optimizer is used with a learning rate of $2e^{-4}$. We report the performance for object detection and instance segmentation in terms of mean average precision (mAP).

\noindent \textbf{ADE20K} \cite{ade} is a challenging segmentation dataset, which contains about 20,000 images and covers 150 categories. We apply the ImageNet pre-trained TinyVim as backbone to extract image features and Semantic FPN \cite{fpn} as decode head for segmentation. Follwing \cite{efficientformer, swiftformer}, we train the segmentation model with an image size of $512 \times 512$ and a batch size of 32 for 40K iterations. The AdamW optimizer with  poly learning rate scheduling is applied and the initial learning rate is $2e^{-4}$. The semantic segmentation performance is reported with the mean intersection over union (mIoU) metric.

\setlength{\tabcolsep}{2pt}
\begin{table}[t]
	\centering
	\caption*{Table A. 
		Ablation for the number of Mamba blocks. Throughput is tested on a Nvidia V100 GPU with maximum power-of-two batch size that fits in memory. 
	}
	\begin{tabular}{c|c|c|c|c}
		\toprule
		Variants &Params&GMACs & Throu. & Top-1 \\
		\hline
		Vim-Ti \cite{vim}&7.0 &1.5 &617 &76.1\\
		EfficientVMamba-T \cite{evmamba}&6.1 &1.0 &1396 &76.5\\
		\rowcolor[gray]{0.92}
		\textbf{TinyViM-S*}& \bf6.2 & \bf0.9 &\bf2235 & \bf78.5\\
		\rowcolor[gray]{0.92}
		\textbf{TinyViM-S}& \bf5.6 & \bf0.9 &\bf2563 & \bf79.2\\
		\bottomrule
	\end{tabular}
	\vspace{-3mm}
	\label{tab:ab}
\end{table}

\section{Model Configuration}
We give the detailed architectural configuration of the TinyViM variant in Table A, which details each building block of the model variant and the corresponding hyperparameters. We mainly place the proposed TinyViM blocks at the end of each stage to capture the global context-enhanced features. For the third stage, we additionally place a TinyViM in the middle of the stage to obtain a better balance between accuracy and efficiency, and the effectiveness of this design of setting up more and larger blocks in the third stage has been confirmed by previous works \cite{onnet,repvit}. In addition, we set the downsampling ratios in TinyViM Blocks of the four stages from small to large to 8, 4, 2 and 1, respectively. In other words, the resolution of the feature maps input to Mamba block at each stage is 1/32 of the original images.
\begin{table*}[t]
	\caption*{Table A. Our detailed model variants for ImageNet-1k, which describe the model configurations at each stage including the numbers of Local Block and TinyVim Block, the output channels and the resolution of the output features. In addition, the expansion factor for FFN in both Local Block and TinyViM Block defaults to 4.} \label{tab:model_variants}
	\begin{center}
		\resizebox{1.0\linewidth}{!}{
			\setlength{\tabcolsep}{2.0pt}{
				\begin{tabular}{ccccc}
					\toprule[1.2pt]
					Name & Output & Small & Base & Large \\
					\midrule[1.1pt]
					stem & 56 $\!\times\!$ 56  & \multicolumn{3}{c}{[RepDW-3 $\times$ 2, stride=2]}\\
					\midrule
					\multirow{1}{*}{stage1} & \multirow{1}{*}{56 $\!\times\!$ 56} 
					& $\left[               
					\begin{array}{cc}
						\!\text{Local Block}\! \!\times\! \text{2}, \text{d=48}  \\ 
						\!\text{TinyViM Block}\! \!\times\! \text{1} \\
					\end{array}
					\right] $  
					&  $\left[               
					\begin{array}{cc}
						\!\text{Local Block}\! \!\times\! \text{3}, \text{d=48}  \\ 
						\!\text{TinyViM Block}\! \!\times\! \text{1} \\
					\end{array}
					\right] $  	
					&  $\left[               
					\begin{array}{cc}
						\!\text{Local Block}\! \!\times\! \text{3}, \text{d=64}  \\ 
						\!\text{TinyViM Block}\! \!\times\! \text{1} \\
					\end{array}
					\right] $  \\		
					\midrule
					
					\multirow{4}{*}{stage2} & \multirow{4}{*}{28 $\!\times\!$ 28} &\multicolumn{3}{c}{[RepDW-3 $\times$ 1, stride=2]} \\
					\cmidrule{3-5}
					&
					&  $\left[               
					\begin{array}{cc}
						\!\text{Local Block}\! \!\times\! \text{2}, \text{d=64}  \\ 
						\!\text{TinyViM Block}\! \!\times\! \text{1} \\
					\end{array}
					\right] $    
					& $\left[               
					\begin{array}{cc}
						\!\text{Local Block}\! \!\times\! \text{2}, \text{d=96}  \\ 
						\!\text{TinyViM Block}\! \!\times\! \text{1} \\
					\end{array}
					\right] $  
					&  $\left[               
					\begin{array}{cc}
						\!\text{Local Block}\! \!\times\! \text{3}, \text{d=128}  \\ 
						\!\text{TinyViM Block}\! \!\times\! \text{1} \\
					\end{array}
					\right] $   \\
					\midrule
					
					\multirow{4}{*}{stage3} & \multirow{4}{*}{14 $\!\times\!$ 14} & \multicolumn{3}{c}{[RepDW-3 $\times$ 1, stride=2]} \\
					\cmidrule{3-5}
					& 
					&  $\left[               
					\begin{array}{cc}
						\!\text{Local Block}\! \!\times\! \text{7}, \text{d=168}  \\ 
						\!\text{TinyViM Block}\! \!\times\! \text{2} \\
					\end{array}
					\right] $   
					&  $\left[               
					\begin{array}{cc}
						\!\text{Local Block}\! \!\times\! \text{8}, \text{d=192}  \\ 
						\!\text{TinyViM Block}\! \!\times\! \text{2} \\
					\end{array}
					\right] $  
					&  $\left[               
					\begin{array}{cc}
						\!\text{Local Block}\! \!\times\! \text{10}, \text{d=384}  \\ 
						\!\text{TinyViM Block}\! \!\times\! \text{2} \\
					\end{array}
					\right] $   \\
					\midrule
					
					\multirow{3}{*}{stage4} & \multirow{3}{*}{7 $\!\times\!$ 7} &\multicolumn{3}{c}{[RepDW-3 $\times$ 1, stride=2]} \\
					\cmidrule{3-5}
					& 
					&  $\left[               
					\begin{array}{cc}
						\!\text{Local Block}\! \!\times\! \text{5}, \text{d=224}  \\ 
						\!\text{TinyViM Block}\! \!\times\! \text{1} \\
					\end{array}
					\right] $    
					& $\left[               
					\begin{array}{cc}
						\!\text{Local Block}\! \!\times\! \text{4}, \text{d=384}  \\ 
						\!\text{TinyViM Block}\! \!\times\! \text{1} \\
					\end{array}
					\right] $  
					&  $\left[               
					\begin{array}{cc}
						\!\text{Local Block}\! \!\times\! \text{5}, \text{d=512}  \\ 
						\!\text{TinyViM Block}\! \!\times\! \text{1} \\
					\end{array}
					\right] $   \\
					\midrule				
					
					Classifier & $1\times1$ & \multicolumn{3}{c}{Average pool, 1000d fully-connected} \\
					\midrule
					\multicolumn{2}{c}{GFLOPs} & 0.9G &  1.5G & 4.7G \\
					\multicolumn{2}{c}{Params} & 5.6M &  11.0M & 31.7M \\
					
					\bottomrule[1.2pt]
				\end{tabular}
			}
		}
		\vspace{4mm}
	\end{center}
\end{table*}

\begin{figure*}[t]
	\centering    \includegraphics[width=0.99\textwidth]{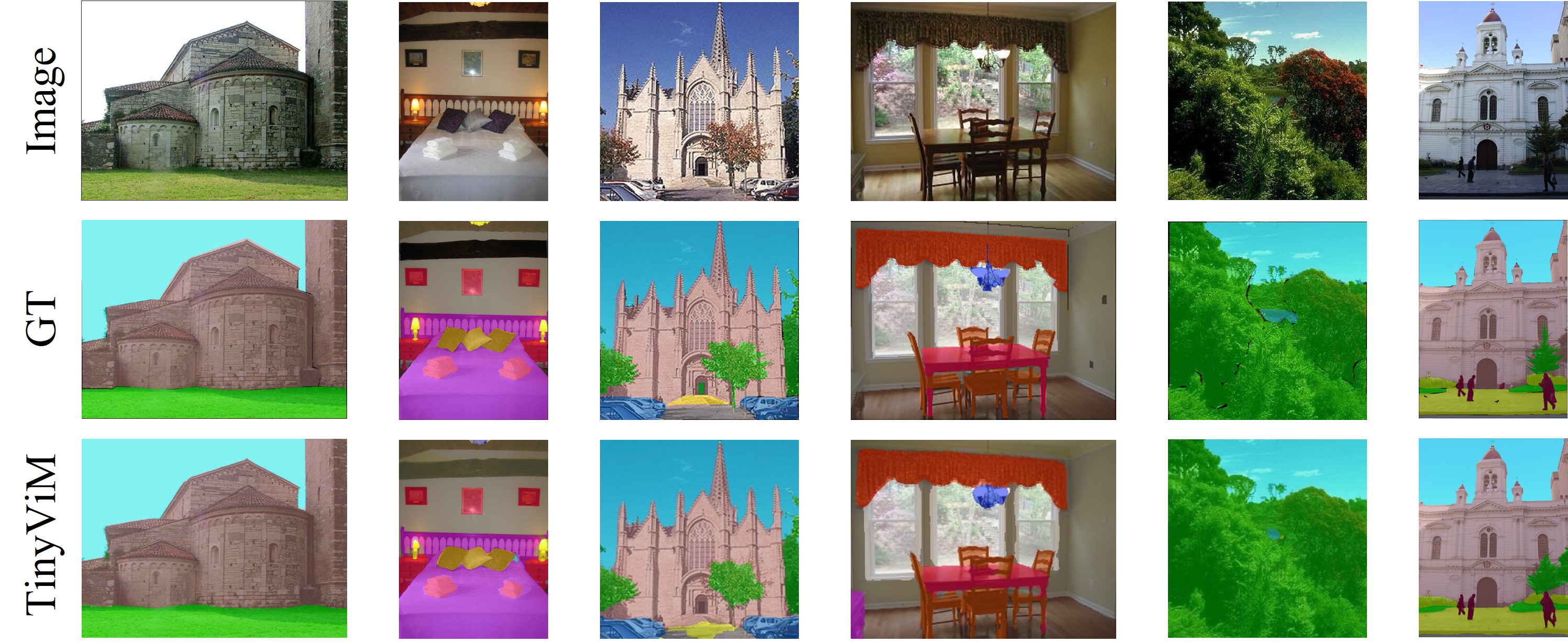}
	\caption*{Figure A. Visualization of the detection and instance segmentation predictions on COCO. It shows that TinyViM can accurately localize and segment various objects in complex scenes.}
	\label{fig:ade}
\end{figure*}

\begin{figure*}[t]
	\centering    \includegraphics[width=0.98\textwidth]{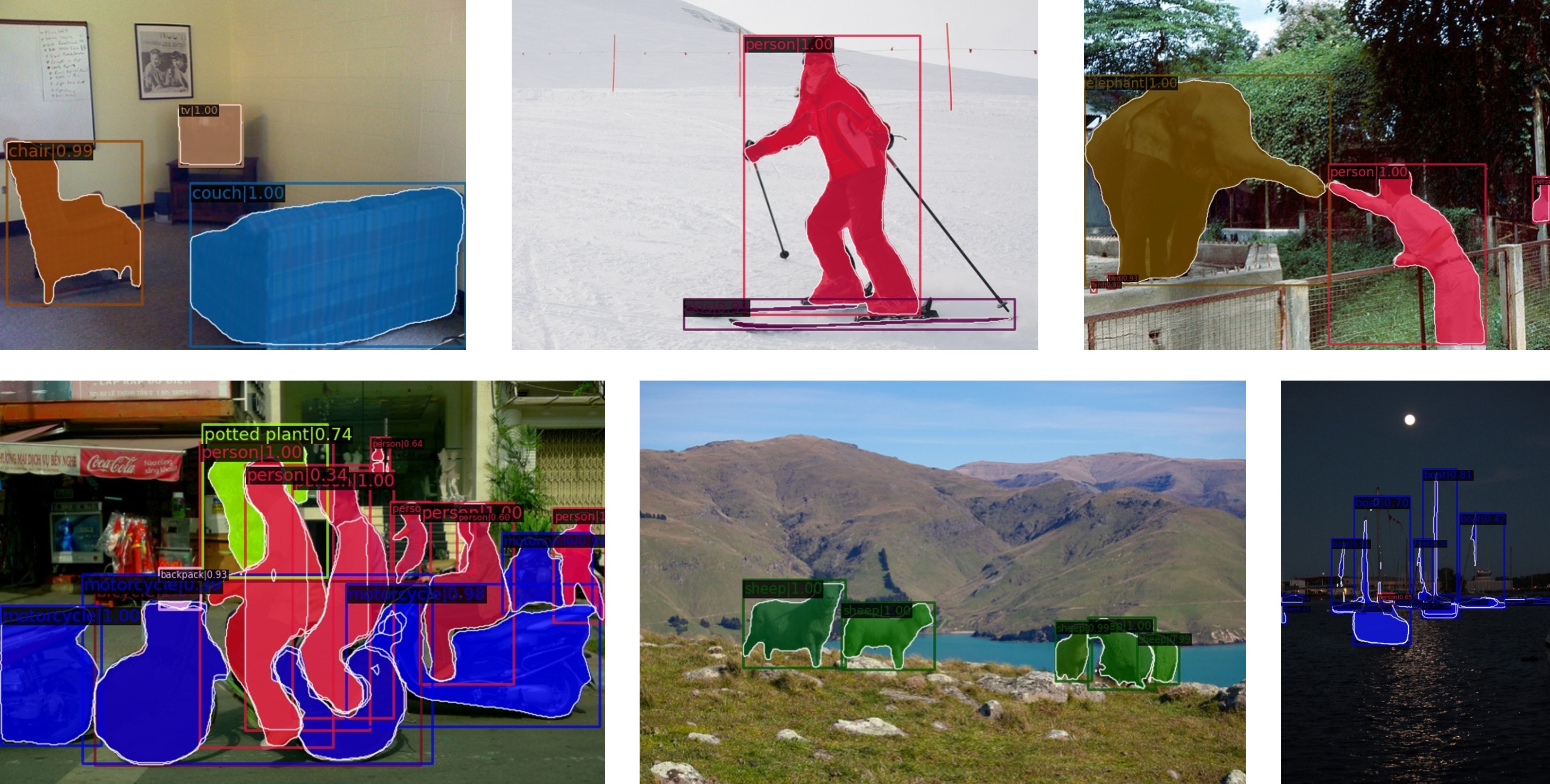}
	\caption*{Figure B. Visualization of segmentation predictions on ADE20K. Top: Input images. Mid: Ground truth masks. Bottom: The semantic segmentation results of TinyViM. It shows that TinyViM can accurately segment various objects in complex scenes.}
	\label{fig:ade}
\end{figure*}
\section{More Comparition Results}
\subsection{Qualitative results on ADE20K}
We visualized the result of segmentation on ADE20K, 053
which is shown in Fig. A. It can be found that the segmen- 054
tation result of TinyViM is very close to the label. The cate- 055
gories and boundaries of objects are accurately segmented. 056
These visualizations demonstrate the state-of-the-art perfor- 057
mance of TinyViM in semantic segmentation

\subsection{Qualitative results on MS-COCO}
The detection and instance segmentation results on coco are 060
visualized in Fig. B. It can be observed that the people 061
and animals in the image are accurately segmented. The 062
location of the detection box is also very accurate. These 063
visualization results show that TinyViM achieves advanced 064
performance in both instance segmentation and detection 065
tasks.

\section{Efficiency Analysis}

We attribute the efficiency of TinyViM to two reasons: 1, the combination of mobile-friendly convolution and 2, the efficient laplace mixer design. For the former, we are not obsessed with designing a pure Mamba architecture such as ViM \cite{vim}. The inductive bias and mobile-friendliness of convolution make it naturally suitable for lightweight backbones. As a result, pure Convolution \cite{mobilenet,mobilenetv2,repvit} or Convolution-ViT \cite{swiftformer,efficientformer,efficientformerv2} models dominate the current lightweight backbone family. Inspired by these excellent Convolutional-ViT hybrid model \cite{swiftformer,efficientformer,efficientformerv2,fastvit}, we design the Convolution-Mamba hybrid architecture TinyViM. For the latter, we propose a variant TinyViM-S* to explore the impact of our reduced number of Mamba blocks on model efficiency, as shown in Table A. Specifically, TinyViM-S* adopts the same architecture as EfficientVMamba-T \cite{evmamba}, i.e., for the four stages, the number of Mamba blocks is (2,2,4,2) and the feature dimensions are (48,96,192,384). It can be observed that TinyViM-S* shows significantly higher throughput compared to EfficientVMamba-T and ViM-Ti. Moreover, the Top-1 accuracy of TinyViM-S* is higher than that of EfficientVMamba-T and VIM-Ti by 2.4\% and \%2.0, respectively. These results validate the effectiveness of Laplace Mixer compared to other Mamba block designs. In addition, the performance and efficiency advantage is further increased when combining convolution at each stage of the network. Therefore, by combining Convolution and Laplace Mixer, TinyViM-S obtains a Top-1 accuracy of 79.2 and outperforms other Convolution, ViT and Mamba based models between accuracy and efficiency.

\end{document}